\documentclass{article}
\usepackage{spconf,amsmath,graphicx,tikz, capt-of}
\usetikzlibrary{positioning}
\usepackage{xcolor,booktabs,dcolumn,caption,enumitem}
\usepackage{adjustbox,url}
\usepackage{hyperref}

\title{Pixel-wise Color Constancy via Smoothness Techniques in Multi-Illuminant Scenes}
\name{Umut Cem Entok, Firas Laakom, Farhad Pakdaman, and Moncef Gabbouj}
\address{Faculty of Information Technology and Communication Sciences, Tampere University, Finland}
\begin{document}
%
\maketitle

\begin{abstract}
\noindent
Most scenes are illuminated by several light sources, where the traditional assumption of uniform illumination is invalid. This issue is ignored in most color constancy methods, primarily due to the complex spatial impact of multiple light sources on the image. Moreover, most existing multi-illuminant methods fail to preserve the smooth change of illumination, which stems from spatial dependencies in natural images. Motivated by this, we propose a novel multi-illuminant color constancy method, by learning pixel-wise illumination maps caused by multiple light sources. The proposed method enforces smoothness within neighboring pixels, by regularizing the training with the total variation loss. Moreover, a bilateral filter is provisioned further to enhance the natural appearance of the estimated images, while preserving the edges. Additionally, we propose a label-smoothing technique that enables the model to generalize well despite the uncertainties in ground truth. Quantitative and qualitative experiments demonstrate that the proposed method outperforms the state-of-the-art.




\end{abstract}
\begin{keywords}
Color Constancy, Multi-Illuminant Estimation, Total Variation, Bilateral Filter, Label-smoothing
\end{keywords}
\section{Introduction}
\label{sec:intro}
Color constancy focuses on eliminating the effects of illumination on the observed colors in captured scenes~\cite{finlayson2001color}. The human visual system is inherently developed to perceive accurate surface colors to some degree, despite varying illuminations causing different color casts~\cite{nikkanen2013computational}. Inspired by this ability, digital cameras aim to counterbalance these color casts in images as if they are captured under canonical illuminant~\cite{ebner2007color}. This process is known as white-balancing, ensuring color constancy.  

In a digital camera, the individual pixel $(x,y)$ of a captured image consists of three elements, spectral power distribution (SPD) of the light source $\phi(x,y,\lambda)$, spectral reflectance of the surface $S(x,y,\lambda)$, and camera sensor spectral sensitivity $C(\lambda)$, all are through the visible spectrum, $w$, based on the wavelength of the light source, $\lambda$~\cite{vonkries}. The image formulation can be expressed as follows:
\begin{align}
  \textbf{I}_{\text{RGB}}(x, y) = \int_{w} \Phi(x, y,\lambda)S(x, y,\lambda)C(\lambda) \,d\lambda ,
\end{align}
\noindent
where $\textbf{I}_{\text{RGB}}(x, y)$ is the RGB image taken under an unknown illuminant. If we assume that the illumination is uniform and single across the entire image, von Kries transformation~\cite{vonkries} holds for white-balancing of the image. This transformation is expressed as:

\begin{align}
    \mathbf{I}_{\text{RGB}}^{\text{a}} &= 
    \mathbf{I}^{\text{a}}
    \cdot \mathbf{I}_{\text{RGB}}^{\text{C}} ,
    \hspace{1em}
    \mathbf{I}^{\text{a}} =
    \begin{bmatrix}
        e_1 & 0 & 0 \\
        0 & e_2 & 0 \\
        0 & 0 & e_3 
    \end{bmatrix},
\end{align}

\noindent
where $\mathbf{I}_{\text{RGB}}^{\text{a}}$ is the image captured under unknown illuminant $a$, $\mathbf{I}_{\text{RGB}}^{\text{C}}$ is the image under canonical illumination, and $\mathbf{I}^{\text{a}}$ is the white-balance gain matrix~\cite{yilmaz2022spectral}. However, uniform and single illumination assumptions are frequently violated in real-world applications, including indoor areas where multiple light sources are present~\cite{dror2004statistical,beigpour2013multi,bianco2017single,li2023mimt}. Therefore, the color casts under the exposition of multiple light sources cannot be corrected with only one von Kries transformation, but a pixel-wise varying transformation~\cite{sidorov2019conditional}. For instance, in the presence of two illuminants, the white-balance gain matrix is modified as follows:
\begin{align}
    \mathbf{I}^{\text{ab}}(x,y) =
    \begin{bmatrix}
        e_1(x,y) & 0 & 0 \\
        0 & e_2(x,y) & 0 \\
        0 & 0 & e_3(x,y) 
    \end{bmatrix},
\end{align}
\noindent
where $\mathbf{I}^{\text{ab}}(x,y)$ is the pixel-wise white-balance gain matrix, referred to as the illumination map in this work, under two illuminants, specifically denoted as $a$ and $b$ light sources. In the context of multi-illuminant color constancy, the goal is to map an input image to a white-balanced image, achieved by estimating the illumination map affected by multiple sources~\cite{sidorov2019conditional}.

There has been many research performed on color constancy in past years~\cite{land1977retinex,hordley2006scene,gijsenij2011computational,fan2022fusion}. With the progress in deep learning methodologies, several learning-based illumination estimation methods have been suggested~\cite{li2023mimt,sidorov2019conditional,hu2017fc4,laakom2019color,laakom2020bag,akazawa2021multi,kim2021large,bianco2017single}. One of the techniques is to divide input images into patches and estimate the illuminants within each patch using Convolutional Neural Networks (CNNs)~\cite{bianco2017single}. While this method stands as the pioneer in employing CNNs for addressing color constancy under multiple illuminants, like most patch-wise models, it does not capture the spatial dependencies among patches and fails to preserve the smoothness of illumination changes in non-overlapping patches. A recent work in~\cite{sidorov2019conditional} presented a Generative Adversarial Network (GAN)-based approach by introducing an additional angular loss term to estimate the illumination map. While this is an innovative solution, the estimated images frequently introduce noticeable spatial artifacts and compromise naturalistic images. The method in ~\cite{kim2021large} constructed a multi-illuminant dataset and proposed U-Net~\cite{ronneberger2015u} for per-pixel white-balancing. Despite the dataset being extensive for multi-illuminant color constancy research, the absence of object reflectivity leads to inaccuracies in the ground-truth model. Another work in~\cite{li2023mimt} focused on the detection of achromatic pixels and surface color similarity to estimate local illuminants and predict the similarity between object colors. However, the overall performance of this approach does not surpass the work in~\cite{kim2021large}.


In this paper, we tackle the challenge of multi-illuminant color constancy by ensuring pixel-wise smoothness with Total Variation~\cite{javanmardi2016unsupervised}. This choice is motivated by the observation that illuminants change smoothly across the image and remain consistent between nearby pixels~\cite{barnard1996colour}; which is missed in most existing works. Furthermore, we utilize a bilateral filter~\cite{choudhury2009color} to retain edges and maintain pixel-wise consistency, thereby improving the overall visual realism of the resulting images. To train and evaluate our multi-illuminant estimation model, we employ the LSMI dataset~\cite{kim2021large}. To mitigate the possible inaccuracies in the ground truth illumination model caused by the lack of surface reflectivity, we propose using a label-smoothing technique~\cite{muller2019does}.



In summary, the main contributions of this paper are as follows:

\begin{itemize}
  \setlength\itemsep{-0.3em}
  \item We introduce a pixel-level learning-based illumination estimation model to address the multi-illuminant color constancy problem.
  \item The method we propose ensures that illuminant changes remain consistent among neighboring pixel locations, achieved through the application of total variation loss during the training of the estimation model.
  \item After the training, we apply bilateral filtering to the estimated image, delivering denoising across the entire image while enhancing photorealism.
  \item To address possible noises in the ground truth illumination model, we propose a label-smoothing approach by introducing Gaussian noise injected into the ground truths.
\end{itemize}


\section{Related Work}
\label{sec:background}

Many earlier methods in color constancy assume that there is one global illuminant in the scene, and the uniformity of colors is preserved. Previous studies~\cite{buchsbaum1980spatial,cepeda2014gray} are based on the average color of the image, gray, which is used to estimate the illuminant. White Patch~\cite{rizzi2002color} and Max-RGB~\cite{land1971lightness} address the earlier statistical-based approaches to the color constancy problem. While these methods do not require a dataset, they do not deliver visually satisfying results if multiple illuminations are present. Recent statistical approaches include convex functions~\cite{abedini2023single}, pixel-level probabilistic model~\cite{laakom2020probabilistic}, and additional mathematical terms by improving Gray-world and Max-RGB assumptions~\cite{finlayson2013corrected,ulucan2022color} to estimate the illuminant of an image. In contrast, learning-based approaches have emerged as leading solutions to color constancy problems recently. Those studies~\cite{bianco2019quasi,lo2021clcc} use CNNs to estimate the illuminant. However, these methods depend on the assumption of uniform color constancy.


Uniformity of illumination is repeatedly violated in real-world settings~\cite{dror2004statistical,li2023mimt,beigpour2013multi}. Therefore, non-uniform illuminant estimation, referred to as multi-illuminant estimation, is examined in recent studies~\cite{gijsenij2011color,joze2013exemplar,beigpour2013multi,hu2017fc4,bianco2017single,sidorov2019conditional,domislovic2021outdoor,kim2021large,akazawa2022n,li2023mimt}. These studies are limited due to the lack of available multi-illuminant datasets. For instance, in~\cite{sidorov2019conditional}, the author proposes the AngularGAN approach to estimate multi-illuminant colors in images whose ground truths are generated by tinting the illumination maps. In ~\cite{kim2021large}, they propose the Large Scale Multi-Illuminant (LSMI) Dataset and a learning-based algorithm. The dataset consists of real-life images illuminated by multiple sources, along with pixel-wise ground truth illumination maps. However, their learning-based method overlooks the absence of object reflectivity, resulting in possible noisy ground truth modeling. Another study in~\cite{akazawa2022n} proposes an N-white balancing to determine white points of illuminants in different segments of the image. A recent work~\cite{li2023mimt} suggests a multi-task learning method. They employ auxiliary tasks to detect achromatic pixels and to predict surface color similarity. Most of these methods evaluate their models with the LSMI dataset. However, they neglect to consider pixel-wise consistent illumination variation.


\section{Proposed Method}
\label{sec:method}

\subsection{Method Overview}
\label{ssec:methodOv}

\begin{figure*}[t]
{\includegraphics[width=19cm]{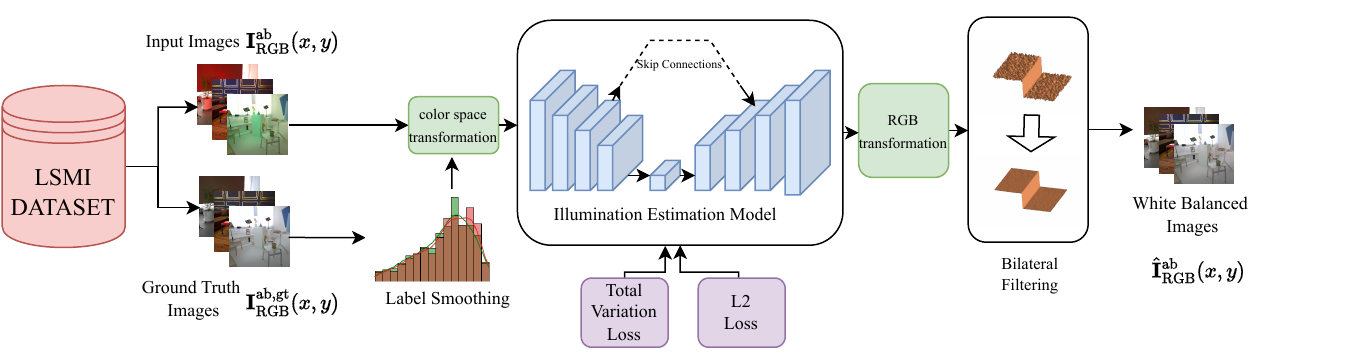}}

\caption{Overview of the proposed method: The U-Net multi-illuminant estimation model processes input images and smoothed ground truth images within the log-chrominance color space. During training, the model learns an illumination mapping between input and ground truth images by minimizing Total Variation Loss and L2 loss. The Bilateral Filter is then applied to the estimated images. The final output is the white-balanced image.}
\label{fig:pipeline}
\end{figure*}

The proposed method models the problem by predicting the pixel-wise illuminants for each input image, through supervised learning. To ensure illuminant smoothness and enhance visual quality over neighboring pixels, we integrate a total variation loss during training and apply bilateral filtering after training. Label smoothing is employed for increasing robustness against noisy ground truth distribution. See Figure~\ref{fig:pipeline} for an overview of the pipeline.

\subsection{Ground Truth Illumination Model}
\label{ssec:illummodel}
Under the assumption that the scene is lit by two illuminants, the ground truth illumination model is defined as follows~\cite{kim2021large}:


\begin{align}
    &\textbf{I}{}_{\text{RGB}}^{\text{ab}}(x,y) = [\alpha(x,y)\cdot l_a + [1-\alpha(x,y)]\cdot l_b]\cdot \textbf{I}_{\text{RGB}}^{\text{ab,gt}}(x,y)
    \\[5pt]
    &\textbf{I}{}_{\text{RGB}}^{\text{ab}}(x,y) =\textbf{\textbf{I}}^{\text{ab}}(x,y)\cdot \textbf{I}_{\text{RGB}}^{\text{ab,gt}}(x,y)   ,
\end{align}

\noindent
where  $l_a$ and $l_b$ are chromaticities of $a$ and $b$ illuminants, $\alpha(x,y)$ is the illumination weight of the illuminant $a$ at pixel $(x,y)$, $\textbf{I}{}_{\text{RGB}}^{\text{ab}}(x,y)$ is the input RGB image, $\textbf{I}_{\text{RGB}}^{\text{ab,gt}}(x,y)$  is the ground truth RGB image, and $\textbf{I}^{\text{ab}}(x,y)$ is the pixel-wise ground truth illumination map. Thus, $\alpha$ defines $\textbf{I}^{\text{ab}}$ that does the white balancing operation to the input image.
\subsection{Color Space Transformation}
\label{ssec:io}
The proposed algorithm takes a linear-raw RGB input image and the corresponding ground truth image. Then it converts them to log-chrominance space. This conversion includes a logarithm that corresponds to the chromaticity of illumination in terms of camera-sensor response. The conversion results in a simpler color space than the RGB triplet although the absolute scale ambiguity is not preserved~\cite{barron2015convolutional}. The log-chrominance space is defined as follows:
\begin{align}
  &u(x,y) = \log\left(\frac{I_{\text{R}}(x,y) +\epsilon}{I_{\text{G}}(x,y) +\epsilon}\right) 
  \\[5pt]
  &v(x,y) = \log\left(\frac{I_{\text{B}}(x,y) +\epsilon}{I_{\text{G}}(x,y) +\epsilon}\right) ,
\end{align}
where $\epsilon$ is an infinitesimal number to avoid division by zero error and $I_R(x,y), I_G(x,y), I_B(x,y)$ are spatial pixels of each R, G, and B channels, respectively. Note that the estimation model utilizes the log-chrominance color space, \textit{uv}, for the input, the ground truth, and the predicted output. Also, the estimated illumination map is transformed back to the RGB color space.
\subsection{Estimation Model}
\label{ssec:model}
The estimation model, based on the U-Net model~\cite{ronneberger2015u} and detailed in~\cite{kim2021large}, serves as the baseline model in this work. It comprises 16 convolutional steps, with 8 dedicated to downsampling and 8 for upsampling. The network takes the input and ground truth images in log-chrominance space and shrinks them to 256x256 image size. Following this, the network proceeds to estimate the illuminants, producing the output image in the same color space.
\subsection{Label Smoothing}
\label{ssec:labelsmooth}

In the context of constructing the ground truth illumination map (referenced in Section~\ref{ssec:illummodel}), the alpha parameter, $\alpha$, plays a crucial role. However, the ground truths in LSMI dataset~\cite{kim2021large} are extracted only considering the reflected light and not the incoming light, thus the surface reflectivity is not accounted for. Therefore, the actual ground truth is considered to be a noisy version of the ground truth in~\cite{kim2021large}. To address this, we propose applying label-smoothing ~\cite{muller2019does}, which not only prevents the network from becoming over-confident to the noisy ground truth but also improves the generalization of estimating illumination maps.

Label smoothing is introduced by adding a random Gaussian noise to the $\alpha$ in every training batch. The distribution has zero mean and standard deviation equal to one-tenth of the magnitude of the corresponding $\alpha_{raw}(x,y)$. That is:
\begin{align}
  &\sigma_n(x,y) = \frac{\alpha_{raw}(x,y)}{w_{n}} \quad w_n = 10
\end{align}
\begin{align}
  &\textit{X}(x,y) \sim \mathcal{N}(0, \sigma_n^2(x,y))
\end{align}
\begin{align}
  &\alpha_{smooth}(x,y) = \alpha_{raw}(x,y) + \textit{X}(x,y) ,
\end{align}

\noindent
where $w_n$ is the smoothing constant, $\sigma_n$ is the standard deviation, $\textit{X}(x,y)$ is the random Gaussian noise, $\alpha_{raw}(x,y)$ is the raw illumination weight, and $\alpha_{smooth}(x,y)$ is the smoothed illumination weight, all evaluated at the $(x,y)$ pixel.

\begin{figure}[h]
\centering
\centerline{\includegraphics[width=5cm]{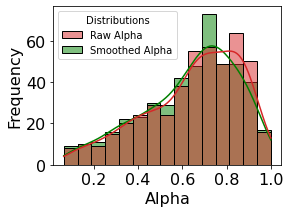}}
\caption{Distributions of $\alpha_{raw}(x,y)$ and $\alpha_{smooth}$ parameters}
\label{fig:labelsmooth}
\end{figure}

\noindent
Figure~\ref{fig:labelsmooth} illustrates the distributions of raw alpha and smoothed alpha values during a single training iteration. The visualization demonstrates that label smoothing transforms the $\alpha$ distribution to closely resemble a Gaussian distribution, indicating a smoother distribution of hard labels.
\subsection{Total Variation Loss}
\label{ssec:loss}
We anticipate that the illumination map, derived from multiple sources, should demonstrate piece-wise smoothness due to the gradual variation of illuminants across the image~\cite{javanmardi2016unsupervised}. To accomplish this, the Total Variation (TV) Loss is proposed. This loss imposes an extra penalty to the overall loss function, ensuring smoothness across illuminant changes. TV Loss is defined as:
\begin{align}
L_{\text{TV}} = \sum_{x,y \in I} | \nabla_X f(x,y) | &+ | \nabla_Y f(x,y) |
\end{align}
\begin{align}
&= \sum_{i,j \in I} |\hat{\textbf{I}}_{\text{RGB}}^{(i,j)} - \hat{\textbf{I}}_{\text{RGB}}^{(i+1,j)}| + \sum_{i,j \in I} |\hat{\textbf{I}}_{\text{RGB}}^{(i,j)} - \hat{\textbf{I}}_{\text{RGB}}^{(i,j+1)}| ,
\end{align}
\noindent
where $\hat{\textbf{I}}{}_{\text{RGB}}^{(i,j)}$ is the estimated RGB value at $(i,j)$.
The overall loss function consisting of a linear combination of L2-loss and TV Loss becomes:
\begin{align}
  &L = L_2 + \lambda_{\text{TV}}\cdot L_{\text{TV}} ,
\end{align}
\noindent
where $\lambda_{\text{TV}}$ is the weight of the TV Loss. Overall, the TV Loss serves to regularize the estimation of the illumination map, accounting for spatial dependencies. 
\subsection{Bilateral Filtering}
\label{ssec:bfilter}

To achieve visually realistic white-balanced images, we propose using a bilateral filter, known for its spatial and range filtering properties~\cite{tomasi1998bilateral}. These properties play a pivotal role in maintaining pixel-wise consistency and preserving edges across the image. The formulation of the bilateral filter is as follows:


\begin{align}
  &BF(\sigma_s,\sigma_r) = \frac{1}{w} \sum_{\mathbf{p},\mathbf{q} \in S} G_{\sigma_s}(\|\mathbf{p} - \mathbf{q}\|) \cdot G_{\sigma_r}(|I_\mathbf{p} - I_\mathbf{q}|) \cdot I_\mathbf{q} .
\end{align}
\noindent
In this equation, $I_\mathbf{p}$ and $I_\mathbf{q}$ represent the intensity at pixel $\mathbf{p}$ and $\mathbf{q}$, and $w$ is a normalization factor to ensure the weights sum up to 1. The parameters $G_{\sigma_s}$ and $G_{\sigma_r}$ are Gaussian functions assuring smoothing and edge-preserving with standard deviations $\sigma_s$ and $\sigma_r$, respectively. Bilateral filtering in the context of color constancy is given as:
\begin{align}
  &\hat{\textbf{I}}^{f} = BF(\sigma_s,\sigma_r)*\hat{\textbf{I}} .
\end{align}

\noindent
Let $\hat{\textbf{I}}$ represent the output RGB image of the estimation model, $BF(\sigma_s, \sigma_r)$ the bilateral filter with parameters $\sigma_s$ and $\sigma_r$ set as 75, and ($*$) denotes a convolution with a 9-pixel diameter. These parameter choices optimize noise filtering and edge preservation.

\section{Experiments and Discussion}
\label{sec:experiments}

The proposed method is trained and assessed on the Large Scale Multi-Illuminant (LSMI) dataset~\cite{kim2021large}, which comprises real-world 7,486 images acquired by three distinct cameras under 1-3 varied illuminants, containing approximately 3000 scenes. Our evaluation focuses exclusively on images captured by the Samsung Galaxy Note 20 Ultra, resulting in a subset of 2360 images illuminated by only two illuminant sources. The resulting dataset is partitioned into training, validation, and test sets with a ratio of 0.75:0.2:0.05, respectively. Note that color checkers in images are masked with black color.

In the training process, the estimation model is trained for 2000 epochs with a batch size of 32. Two different learning rates, namely $0.0001$ and $0.0005$, are determined based on the weight of the TV Loss. These two settings give rise to two distinct proposed models, referred to as Pixel-wise Color Constancy (PWCC), PWCC\_v1 and PWCC\_v2. Those parameters and various settings can be investigated in table~\ref{table:ours}. Moreover, the learning rates exhibit a decay factor of 800, beginning from the 800th epoch. The training is performed with 4 NVIDIA GeForce RTX 2080 Ti GPUs.
\setlength{\textfloatsep}{0.2cm}
\begin{table}[!h]
\small
\caption{Different settings for the Estimation Model of our method on Two-Illuminant images of LSMI dataset~\cite{kim2021large}.}
\begin{tabular}{l|c|c|c|c}
\toprule
\textbf{Method} & Filtering & Label Smooth & $\lambda_{\text{TV}}$ & Learning Rate \\
\midrule
\midrule
PWCC\_v1                         & Yes & No & 2e-4 & 5e-4 \\
PWCC\_v2                         & Yes & Yes & 2e-3 & 1e-4 \\
\bottomrule
\end{tabular}
\label{table:ours}
\end{table}

\begin{figure*}[t]
\centering
{\includegraphics[width=18cm]{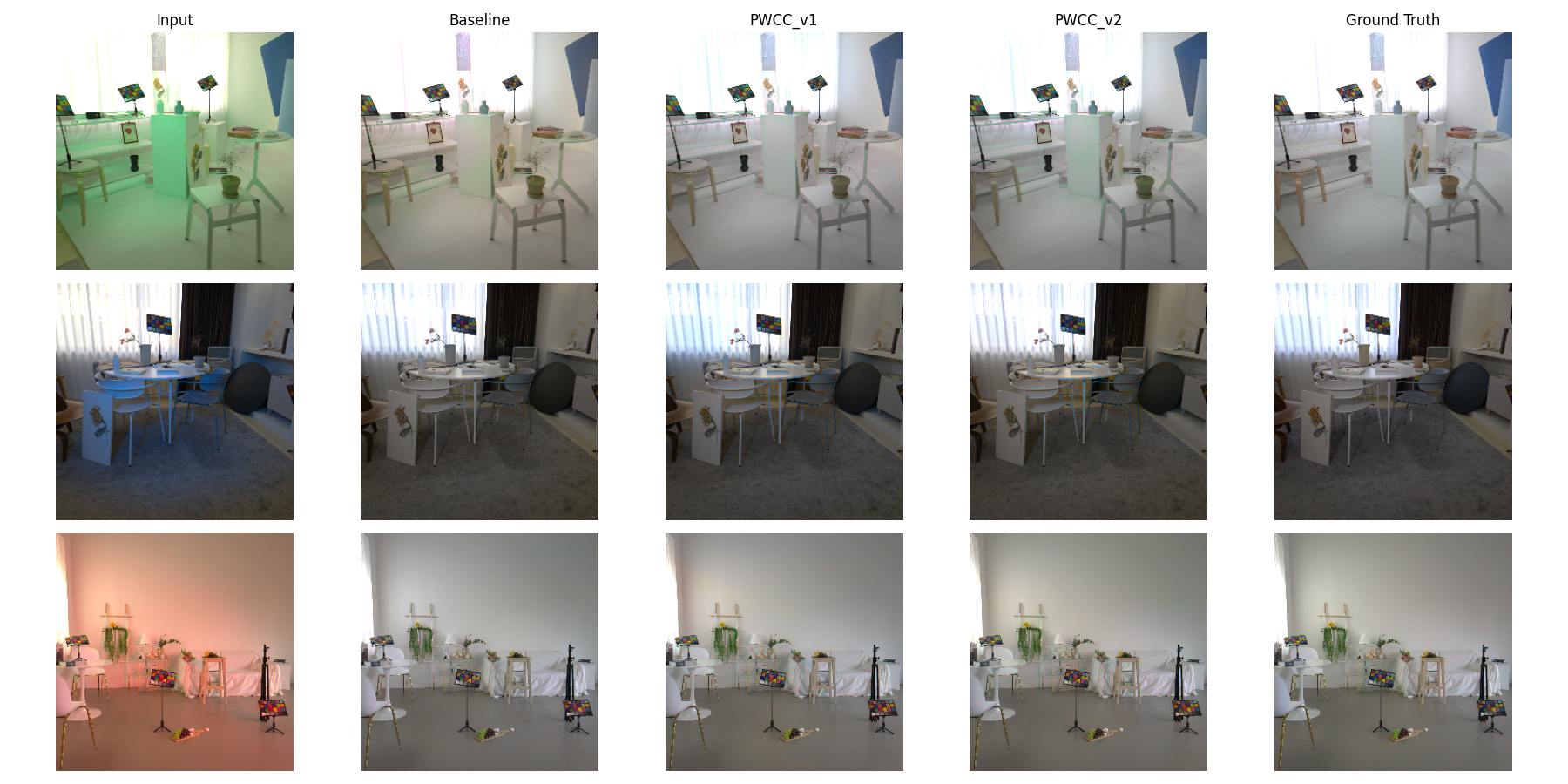}}
\caption{Visualization results on different two-illuminant images of LSMI Dataset~\cite{kim2021large}. From left to right, input image, baseline (LSMI-U-Net~\cite{kim2021large}), PWCC\_v1, PWCC\_v2 predictions, and the ground truth image, respectively.}\label{fig:visual}
\label{fig:qualitative}
\end{figure*}





\subsection{Evaluation Metrics}

The proposed approach is evaluated using a total of 112 two illuminant test images including corresponding ground truth images. Evaluation is quantitatively done by calculating recovery angular errors between ground truth and predicted illumination maps, as it is an intensity-independent error measure~\cite{hordley2004re} and is commonly used in evaluating the performances of color constancy algorithms.  For qualitative comparison, predicted images of our best two models along with the input and ground truth images are compared with the prediction of the baseline method, that is, LSMI-U-Net described in~\cite{kim2021large}.
\\[5pt]
\noindent
\textbf{Quantitative Evaluation} We report the mean, median, worst 25\%, and best 25\% of angular recovery error, $e_{recovery}$, between the ground truth and the estimated illumination maps. $e_{recovery}$ is defined as follows:

 
\begin{align}
  &e_{\text{recovery}}(\textbf{I}^{\text{ab}}, \hat{\textbf{I}}{}^{\text{ab}}) = \cos^{-1} \frac{\textbf{I}^{\text{ab}} \cdot \hat{\textbf{I}}{}^{\text{ab}}}{\|\textbf{I}^{\text{ab}}\| \: \|\hat{\textbf{I}}{}^{\text{ab}}\|} ,
\end{align}

\noindent
where $\textbf{I}^{\text{ab}}$ and $\hat{\textbf{I}}{}^{\text{ab}}$ are the ground truth and the estimated illumination maps for an image, and ($\cdot$)  is the dot product. $e_{\text{recovery}}(\textbf{I}^{\text{ab}}, \hat{\textbf{I}}{}^{\text{ab}})$ is calculated for each image and used for the above-mentioned reports.
\setlength{\textfloatsep}{0.2cm}
\begin{table}[ht]

\centering
\caption{Recovery angular errors on Multi-Illuminant images of LSMI dataset~\cite{kim2021large}}
\begin{tabular}{l|c|c|c|c}
\toprule
\textbf{Method} & \textbf{Mean} & \textbf{Median} & \textbf{W.25\%} & \textbf{B.25\%} \\
\midrule
\midrule
Gray World~\cite{cepeda2014gray}  & 11.3 & 8.8 & 20.74 & 4.93 \\
White Patch~\cite{rizzi2002color} & 12.8 & 14.3 & 23.49 & 5.6 \\
LSMI U-Net~\cite{kim2021large}    & 2.31 & 1.91 & 4.24 & 1.01\\
PWCC\_v1                           & 2.08 & 1.74 & \textbf{3.8} & 0.9 \\
PWCC\_v2                           & \textbf{2} & \textbf{1.7} & \textbf{3.8} & \textbf{0.86} \\
\bottomrule
\end{tabular}
\label{table:quan}
\end{table}

\noindent
Table~\ref{table:quan} reports the recovery angular errors obtained from various methods. We compare our proposed methods with Gray-World~\cite{cepeda2014gray}, White Patch~\cite{rizzi2002color}, and the reproduced results of the state-of-the-art multi-illuminant method LSMI U-Net~\cite{kim2021large}. 
\noindent
It is observed that our proposed methods, PWCC, outperform the competing methods. Specifically, PWCC\_v2 achieves an impressive mean error of 2, which is a 13\% improvement over the best-competing method. A notable observation is that not only the proposed method improve mean performance, but it also significantly improves the results of the worst 25\%, which are often the corner cases.


\newpage
\noindent
\textbf{Qualitative Evaluation} Figure~\ref{fig:qualitative} showcases three visual examples, from the proposed methods and the LSMI-U-Net as the baseline. It can be observed that our method consistently outperforms the baseline, demonstrating better white balancing across various input images. For instance, in the top row image, the baseline cannot fully remove the greenish color cast, while the proposed shows a much better result. Moreover, for the third-row image, the baseline overcompensates for the reddish color cast, which also removes the natural color on the wall. However, in some test data instances, such as the second row, the two methods show competing performances.

\section{Conclusion}
\label{sec:conclusion}
In this paper, a multi-illuminant pixel-wise color constancy approach is proposed. We highlighted the importance of smooth illumination changes in natural images and proposed total variation loss for regularization during training. We proposed to filter the output white-balanced images of the estimation model with a Bilateral Filter, to guarantee the pixel-wise consistency across the image and create a visually natural appearance. Moreover,  the usage of the label smoothing technique ensures that the model generalizes better on the noisy ground truth data. Experimental results demonstrate that the proposed method outperforms the current state-of-the-art method. Future work includes proposing an estimation model accounting for surface reflectivity to enhance the accuracy of ground truth modeling for the multi-illuminant dataset used in this paper.
\section{ACKNOWLEDGEMENT}
This project has received funding from the European Union’s Horizon 2020 research and innovation program under the Marie Skłodowska-Curie grant agreement No [101022466], and the NSF-Business Finland Center for Big Learning (CBL), Advanced Machine Learning for Industrial Applications (AMaLIA) under Grant 97/31/2023.

\bibliographystyle{IEEEbib}
\bibliography{strings}
\end{document}